# Getting More Out Of Syntax with PROPS

**Gabriel Stanovsky, Jessica Ficler, Ido Dagan, Yoav Goldberg**
Computer Science Department, Bar-Ilan University
{gabriel.satanovsky,jessica.ficler,yoav.goldberg}@gmail.com
dagan@cs.biu.ac.il

## Abstract

Semantic NLP applications often rely on dependency trees to recognize major elements of the proposition structure of sentences. Yet, while much semantic structure is indeed expressed by syntax, many phenomena are not easily read out of dependency trees, often leading to further ad-hoc heuristic post-processing or to information loss. To directly address the needs of semantic applications, we present PROPS – an output representation designed to explicitly and uniformly express much of the proposition structure which is implied from syntax, and an associated tool for extracting it from dependency trees.

## 1 Introduction

Propositions, statements for which a truth value can be assigned (e.g., "*Bob loves Mary*"), constitute the primary unit of information conveyed in texts. Accordingly, recognizing and matching proposition structure is a central component in systems and algorithms that attempt to extract semantic information from text, such as question answering, summarization, or recognizing textual entailment, collectively referred as semantic applications.

Several practices exist for recovering the proposition structure of sentences. *Dependency trees* (de Marneffe and Manning, 2008b) are attractive as they directly connect verbal predicates to their arguments, while deep syntax extensions of dependency trees also mark long distance dependencies, further broadening predicate-argument coverage (Ballesteros et al., 2014). Other annotations choose a more semantic approach. Notable examples are *Semantic Role Labeling* (SRL) (Carreras and Màrquez, 2005a), which extracts frames linking predicates with their semantic arguments, or, more recently, *Abstract Meaning Representation* (AMR) (Banarescu et al., 2013), which aims to extract a graph representation capturing the semantic structure of the sentence.

Currently, dependency trees are perhaps the most commonly used representation, mainly for the high accuracy with which they can be obtained, in contrast to the relatively lower performance for current SRL systems and AMR parsers.

Despite the attractive properties of (deep) dependency parses, it is quite hard to read out from them the *complete* structure of *all* propositions expressed in a sentence, for several reasons: (1) Different predications are represented in a non-uniform manner (e.g, passive vs. active, verbal vs. adjectival predication, appositions vs. copula) (2) Proposition boundaries are not easy to detect; and (3) Substantial parts of the dependency structure represent syntactic detail that is not core to proposition structure.
For these reasons, it is common for NLP systems to tailor sets of rules and heuristics to unify and extract specific information from the parse structure. While these heuristics are created mostly in an ad-hoc manner and differ from application to application (e.g., (Culotta and Sorensen, 2004)), the possibility of creating such heuristics rests on rather strong linguistic grounds – the syntactic structure reliably expresses a wide range of predications. This stands in contrast with the deeper analysis required in order to obtain semantic annotations such as AMR or SRL, which are not readily attainable from syntax.

Our goal in this work is to alleviate the need for tailoring such heuristic post-processing of parse trees on a case by case basis. We do this by provid-ing a convenient output format which directly represents the proposition structure that can be induced from syntax, while following several desired design principles. In particular, we uniformly represent propositions headed by different types of predicates, verbal or not. We canonicalize different syntactic constructions that correspond to the same proposition structure, and decouple independent propositions while clearly marking proposition boundaries. Finally, the scheme masks non-core syntactic detail, yielding cleaner compact structures. Overall, we enable simple access to the represented propositions by a uniform graph traversal, offering an appealing alternative input to common uses of dependency structures.

We bundle an implementation of these principles in PROPS (*Proposition Structure*) – an automatic converter which takes as input Stanford dependency trees, and outputs proposition structures with our notation. In most cases. PROPS stays close to the syntactic level, keeping the relatively high accuracy and robustness levels currently attainable by syntactic parsing, while still providing a significant improvement over the bare syntactic structure for downstream semantic processing. In some other cases, where information is currently not reliably produced by automatic parsers (such as difficult conjunctions or differentiating between raising and control), PROPS applies a set of heuristics to obtain the output structure. We demonstrate the out-of-the-box attractiveness of PROPS on a reading comprehension task.

In addition to providing an automatic tool to convert dependency trees we also provide a large semi-automatic annotation of the WSJ portion of the Penn Tree Bank (PTB) (Bies et al., 1995). This recovers our structures with higher accuracy, and includes an accurate handling for the previously mentioned semantically hard decisions. We do so by utilizing gold phrase-based annotations and other annotated resources, such as Propbank (Kingsbury and Palmer, 2003) or Vadas and Curran's (2007) NP structure annotation. We intrinsically test the quality of our predicted structures over the PTB by measuring their accuracy against a small manually annotated subset.

By identifying and clearly enumerating the typical properties that semantic applications would need from syntactic structures, we hope to trigger research on an application-oriented syntactic middle-ground, as opposed to current mostly ad-hoc and under-documented post-processing solutions.

## 2 Design Principles

We focus our output representation around the proposition structure needed for semantic applications, instead of the syntactic focus of dependency trees. To that end, we suggest five desired structural principles that we find missing or lacking in dependency trees. This section presents these principles, as well as consequent formulation of the PROPS output format. The next two sections describe the corresponding dependency tree conversions implemented in PROPS. Section 3 describes the bigger bulk of conversions. These can be realized with high precision in a principled way from dependency trees attainable from current dependency parsers. Few cases, in which we applied heuristics to address some harder decisions, are described in Section 4.

**Masking non-core syntactic detail (Section 3.1)** We want to focus the application on the semantic components of the sentence. While dependency trees are often over specified, and include a node for every token, we would like to remove auxiliary words and instead encode their syntactic function as features in a canonicalized way. Additionally, we would like to group atomic units (such as noun compounds) within a single node.

**Representing propositions in a uniform manner (Section 3.2)** We would like the application to uniformly access a wide range of propositions, headed by different types of predicates, unlike the verb-centric representation of input dependency trees.

**Canonicalizing and differentiating syntactic constructions (Sections 3.3, 4.1)** To ease the semantic handling for downstream applications, we want our structures to (1) Unify the representation of propositions which are semantically equivalent, yet look different in dependency trees, due to syntactic subtleties, and (2) Differentiate syntactically-similar, yet semantically-different, constructions.

**Marking proposition boundaries (Section 3.4)** Clearly marking the minimal span of a standalone proposition and its elements (predicate, arguments, and modifiers) is beneficial for semantic tasks (e.g.,

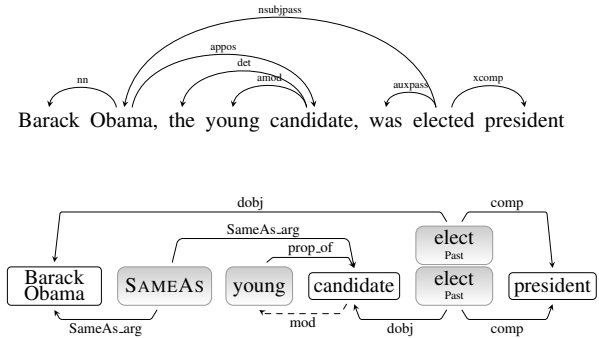

Figure 1: PROPS structure (bottom, detailed in Sections 3 and 4) vs. dependency representation (top) for the sentence "*Barack Obama, the young candidate, was elected president*"

(Angeli et al., 2015)), and not trivially available in depdendency trees.

**Propagating Relations (Section 4.2)** We would like that every relation which is inferable through parse tree traversal (for instance, through conjunctions) would be explicitly marked in our representation. This way we can save an application the need to perform subsequent passes and propagations over the input representation.

### 2.1 Output Format

To achieve our desired principles, we choose a representation format that resembles dependency structures, with the following changes:

**Typed nodes** In order to clearly identify the propositions in our structures, we differentiate between two types of nodes (compared with syntactic dependencies, where there is only one type of nodes): (1) Predicates, which evoke a proposition and (2) Non-predicates, which can be either arguments or modifiers. Additionally, we delegate function words, such as modality and tense, to features of nodes.

Figure 1 depicts the sentence "*Barack Obama, the young candidate, was elected president*" in dependency representation (top) verus PROPS output (bottom). PROPS clearly marks the predicates **elect**, **young** and **SameAs** (representing apposition) as shaded nodes in the graph, along with their features (for instance, past tense indicated in subscript for the predicate **elect**), and their direct arguments and modifiers.

| Relation | Examples |
|---|---|
| *subj* - subject | 1, 2, 5 |
| *dobj* - direct object | 2, 5, 6 |
| *iobj* - indirect object | |
| *comp* - complement | 1, 4 |
| *prep* - preposition | 3,4 |
| *time* - temporal expressions | 8 |
| *prop_of* - adjectival predication | 1,2,6 |
| *SameAs_arg* - argument of SameAs | 3 |
| *outcome* -main clause of conditional | 5 |
| *condition* -dependent clause of conditional | 5 |
| *mod* - modifier | 2, 3, 6 |
| *source* - modal-like modification | 8 |
| *poss* - possessive | |
| *conj* - element in conjunction | 14 |

Table 1: Label set definition and examples index for predicate-argument relations (top) and head-modifier relations (bottom)

**Breaking the correspondence between nodes and words** We simplify the graph structure by allowing multi-word nodes (e.g., Barack Obama), versus having each node corresponding to a single word in dependency trees. In some cases (as in the **SameAs** node), nodes do not correspond to specific words in the sentence (see Section 3.2).

**Graph Structure** Similar to the deep variant of dependency trees, our resulting structures are no longer limited to trees. Instead, our structures are directed graphs, as seen in Figure 1.

**Focused edge label set** In order to further simplify the reading of the graph, we introduce a label set of 14 relations (compared with approximately 50 in Stanford dependencies). These enable the user to focus on a more general set of relations between the proposition elements. See table 1 for the complete inventory, along with an index to examples in the paper.

## 3 The PROPS Converter

In this section we describe the transformations carried out in PROPS, a rule-based converter of Stanford dependency-trees.[1] These transformations fulfill the guiding principles described in the previous section. Specifically, we target phenomena which we find to be both feasibly attainable from dependency trees, as well as common enough[2] to be of practical significance for semantic applications, as demonstrated in Section 7.1.

---

[1] We will make the code available upon acceptance.
[2] As we identified by frequency analyses over the PTB.

### 3.1 Masking Non-Core Syntactic Detail

In addition to single word nodes, we produce structures containing multi-word nodes, for cases of multi-word predicates (e.g., **take a picture**) or noun compounds (e.g., New York). In certain syntactic formalisms, these constructs are represented as multiple nodes joined by some designated label. In particular, PROPS relies on the nn and mwe relations in Stanford dependencies for noun phrases and multi-word expressions, respectively. Joining such entities in a single node reduces the size of the tree, reduces the label set, and overall simplifies its processing.

Additionally, we write words in their base form and encode features for modality, negation, definiteness, tense, and passive or active voice for each node in a flat key=value structure (e.g., "*was elected*" appears as a single **elect** node with a past tense feature in Figure 1). Recovering such properties can be done with high accuracy based on lexical and syntactic cues in a few deterministic rules (see Section 7.1 for an intrinsic evaluation). These feature encodings provide abstraction over the different ways in which the features are realized in the surface forms. In addition, we unify markers which are not traditionally thought of as tense. For example, we reduce constructions such as **going to + verb** to indicate future tense (e.g., in "*I am **going to** dance*").

### 3.2 Representing Propositions in a Uniform Manner

Beyond verbal predicates, we deal with three common types of predications which do not adhere to the simple verbal-oriented representation in dependency trees. We re-arrange the tree structure such that each predicate heads its arguments and modifiers.[3]

**Adjectival predication** While syntactic analyses treat adjectives as modifiers, we explicitly represent also their propositional meaning (for example sentences such as "*the boy is tall*" and "*the tall boy*" evoke the proposition tall(boy)). Adjectival propositions are connected using the "prop_of" relation, denoting the 1-ary predication. We also retain the modification status of an adjective using the "mod" rela-

---

[3]The rather complex case of nominal predication is left to be integrated in future work.

tion, which we will elaborate on in Section 3.3. This results in the following structures:

(1) She said that the boy is tall   (2) She saw a tall boy

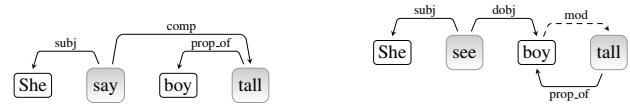

Notice that in (2) we mark "*tall*" as a modifier of "*boy*" for the completeness of the see(she,tall boy) proposition, but "*tall*" is also used as a predicate in a separate tall(boy) proposition.

**Non-lexical propositions** Some propositions do not correspond to a specific word in the sentence, but are rather implied from the syntactic structure. We introduce handling for two such recurring examples: appositions and existentials.

An appositive construction, like "*Barack Obama, the U.S. president, lives in Washington*", implies that "*Barack Obama **is** the U.S. president*", without a word mediating this relation in the sentence. We deal with such cases by introducing a synthetic SAMEAS node for apposition (and copular) constructions. Similarly, we introduce an EXISTS node for existentials (example 4).

(3) Barack Obama, the U.S. president, lives in Washington

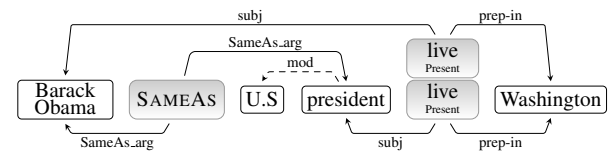

(4) She says that there are signs for rain

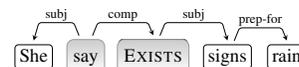

**Conditionals** Conditionals are a syntactic construction in which a word (syntactically referred as "marker") introduces a logical relation between two discrete propositions, where one of the clauses conditions the other. While there has been extensive linguistic study of the subject (Stalnaker, 1981; Bennett, 2003), NLP applications tend to overlook these constructions (a recent exception is (Berant et al., 2014)).

We treat conditional constructions as propositions, as they can be assigned a truth value. We consider the marker word (e.g, "*because*", "*although*",

"*unless*", etc.) as a predicate, taking as argument both clauses, using the "condition" and "outcome" labels, as it introduces the logical relation between them (example 5).

(5) (a) If you build it, they will come
(b) They will come, if you build it

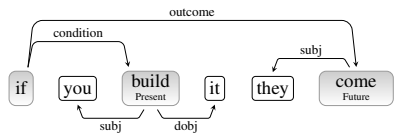

### 3.3 Canonicalizing and Differentiating Syntactic Constructions

Different syntactic realizations may evoke the same proposition structure (e.g., passive versus active voice), while sentences with similarly looking syntactic structures may evoke different proposition structures. We aim to present a unified proposition structure when applicable, and different structures when needed. The major cases we consider are detailed below.

**Adjectival modification** We unify restrictive adjectival modifications that are expressed as prenominal adjectives ("*broken pipe*") or as relative clauses ("*pipe which was broken*"), representing both using the "mod" relation.

(6) (a) The janitor didn't fix the broken pipe
(b) The janitor did not fix the pipe which was broken

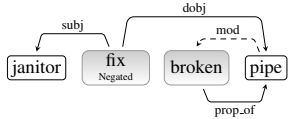

**Adjectival predication** We provide a unified representation for adjectival predications, which are represented using a "prop_of" label connecting a predicate node to its argument. This structure is evoked by adjectives ("*John is nice*") and adjectival phrases ("*John is a nice man*") in the non-restrictive cases of prenominal adjectives, relative clauses, appositive and copular constructions.

(7)
(a) John, who is a nice man
(b) John is a nice man
(c) John, a nice man

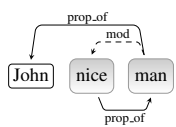

In a more subtle canonicalization, we follow Huddleston et al (2002), and unify, under the above representation, *adjectival complements* as also evoking an adjectival predication.

(8)
(a) You looked very impatient yesterday
(b) You are impatient

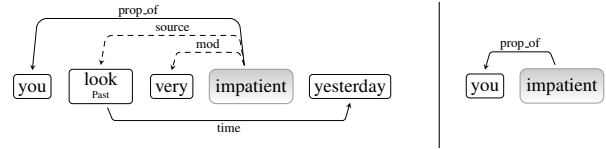

Adjectival complements occur where an adjective appears as a complement of a verbal predicate. For instance, in the sentence: "*you look **impatient***". Dependency representation marks the verb (**look**) as the head of the sentence, and the adjective (**impatient**) as depending on it.

Instead, in PROPS we mark the adjective as the predicate, and the verb as a modifier. We introduce a "source" edge indicating this modification relation. This analysis focuses the attention on the main predicate and maintains a uniform predication structure, as demonstrated by the structural resemblance to "*You are impatient*" (example 8).[4]

Adjectival complements are similar to raising-to-subject constructions ("*you **seem** to be impatient*"). Indeed, we make an attempt to represent both cases in a similar manner, having **impatient** as the main predicate and **seem** as "source". Unfortunately, distinguishing raising from control constructions is more nuanced, and is not handled by current syntactic parers. We handle this heuristically in PROPS, as described in Section 4.1.

**Copulas and appositions** We would like to canonicalize the semantically similar cases of copulas and appositions. Syntactic copular ("*X is Y*") and appositive ("*X, Y*") constructions may both evoke either an equivalence relation ("*Obama, the president*"; "*Obama is the president*") or class - subclass / membership relation ("*Obama, an american citizen*").

The different cases are distinguished based on the syntactic categories and definiteness status of both X and Y, regardless of whether an apposition or a copula appeared in the surface form.

---

[4] While one may think of modality, and hence "source", as being a feature of a predicate rather than a relation, the relation is needed for cases where the modality itself has a rich internal structure, as seen in example (8).

We represent the equivalence case using the SAMEAS predication node (example (3) above) and the membership case as an adjectival predication (as in example (7)).[5]

### 3.4 Marking Proposition Boundaries

While syntactic dependency representations contain much of the proposition structure, they do not clearly mark the boundaries of different propositions and their arguments, making it hard to focus on individual propositions. We use restrictive versus non-restrictive modification to properly bound the scope of arguments (Kamp and Reyle, 1993).

Consider, for instance, the following sentences:
-"*The director who edited 'Rear Window' released Psycho*"
-"*Hitchcock, who edited 'Rear Window', released Psycho*".

While syntactic analysis will assign similar structure for both instances, each induces different proposition boundaries. In the first sentence, the director who edited Rear-Window specifies the minimal scope of a single argument, yielding the proposition released(the director who edited Rear Window, Psycho). However, in the second sentence we can separate the relative clause from the entity it modifies, yielding two distinct propositions: edited(Hitchcock, Rear Window) and released(Hitchcock, Psycho).

The difference between the two cases is that the first exhibits a *restrictive* while the second exhibits a *non-restrictive* modification. Restrictive modifiers are used to select an item from a set and are an integral part of the proposition. We represent these with a "mod" edge (example 9). Non-restrictive modification, on the other hand, "breaks" the propositions and provides additional information on a selected entity, therefore forming a separate predication (example 10).

Similar to the differentiation we make in copulas and appositions, we identify restrictive modification by definiteness status of the entity being modified.

(9) the director who edited 'Rear Window' released Psycho

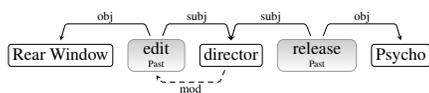

---
[5]The "prop_of" relation is used for both adjectival complements ("*John is beautiful*") as well as noun complements ("*John is a citizen*").

(10) Hitchcock, who edited 'Rear Window', released Psycho

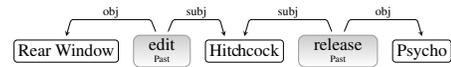

## 4 Heuristically Dealing with Harder Cases

One of the guiding principles for choosing the phenomena described in Section 3 is the ability to provide principled rules for handling them based on automatic parsing output. Yet, certain phenomena are difficult to directly identify from the output of current parsers, and fall out of this scope. However, we find that by applying a few heuristics we can target them with relatively high accuracy, and provide downstream semantic applications with a more practical representation.

While not currently handled by syntactic parsers, most of these distinctions are annotated in existing linguistic datasets. In Section 6 we make use of these annotations and thus are able to provide a gold-standard corpus that covers these phenomena reliably.

### 4.1 Differentiating Raising-to-Subject from Control

Following Huddleston et al (2002)'s analysis, we would like to canonicalize syntactically-different sentences like:

(11) 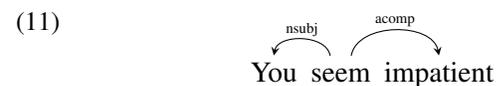

(12) 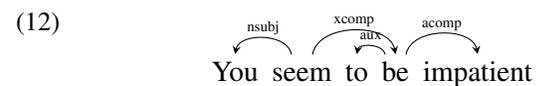

Ideally, in both cases we would like the complement **impatient** to become the main predicate, and mark the verb **seem** as its modifier, as we previously analyzed for adjectival complements (example 8). However, using the output of syntactic parsers it is hard to differentiate example 12 from cases such as example 13, in which we would like to keep the nested dependency representation.

(13) 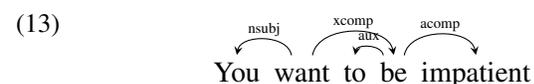

Linguistic theories identify example 12 as *raising-to-subject*, differentiating them from *con-*

*trol* constructions (e.g., example 13), by the correspondence between syntactic and semantic arguments (Davies and Dubinsky, 2008) (i.e., in example 13 "*you*" is a syntactic and a semantic argument of **want**, while in 12 "*you*" is only a syntactic argument of **seem**). However, current syntactic parsers and annotations standards do not make this distinction.

In order to identify raising-to-subject constructions in PROPS, we heuristically use a set of approximately 30 verbs which were found by (Chrupała and van Genabith, 2007) to frequently occur in raising constructions (including verbs such as "seems", "looks", "remains", etc.). For these verbs, similarly to the handling of adjectival complements, we do not produce a proposition, but instead use a "source" edge indicating a modification relation.

### 4.2 Propagating Relations

In some cases, most notably in coordination constructions, predicate-argument relations can be propagated according to the syntactic structure, resulting in new propositions. For example in "*Dell makes and distributes products*" the propagation results in the three propositions make(Dell, products), distribute(Dell, products) and make_and_distribute(Dell, products). Similarly, for "*Dell sells laptops and servers*" we get sell(Dell, laptos), sell(Dell, servers) and sell(Dell, laptops and servers).

Our representation of coordination follows the Prague-Treebank style (Popel et al., 2013; Böhmová et al., 2003) and posits the coordinating word as the main node of the coordinating-conjunction structure, while the different conjuncts are attached to it using "conj" edges.

(14) Kim and Pat speak Spanish / Kim and Pat are a couple

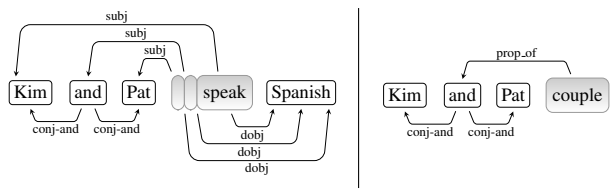

The semantic of coordination nodes is that the node represents the entire conjunction. Following relation propagation, relations may involve the coordination node (i.e. the entire coordinated structure) and/or the individual conjuncts.

A similar approach is taken by the "propagated" variant of Stanford Dependencies (de Marneffe and Manning, 2008a), but we differ in choosing the coordination as the main element in the conjunction, and take more care in distinguishing different kinds of coordination constructions and determining the conjunction scope. As one example, we distinguish between distributive and joint coordination (Huddleston et al., 2002). In distributive coordination ("*Kim and Pat speak Spanish*") we fully propagate the relations, while for cases of joint coordination ("*Kim and Pat are a couple*")[6] we do not propagate, as the relation is meant only for the combination of both entities. When the relation propagation results in multiple propositions involving the same predicate, the predicate node is duplicated in order to distinguish the different propositions and keep with the 1-1 correspondence between propositions and predicate nodes.

We also propagate relations in appositive and copular constructions involving the SAMEAS node (example 3), e.g. "*Amazon, the retail giant, sells products*" will evoke both sell(Amazon, products) and sell(retail giant, products).

### 4.3 Syntactic assertedness

Certain propositions are asserted by the sentence, while others are attributed (compare "*John passed the test*" versus "*the teacher said that John passed the test*"). This property was extensibility studied in the PDTB (Prasad et al., 2006) and PARC (Pareti, 2012) corpora, and was shown to be useful in semantic applications (e.g., (Aramaki et al., 2009)).

Using a dedicated node feature, we mark cases where it is possible to syntactically determine if a proposition is asserted. While hierarchical structure generally implies dependence of the nested proposition, in certain constructions the nested proposition is in fact independent of its subordinating clause. This is the case for non restrictive relative clause ("*Alfred Hitchcock, who directed Psycho*" in which the **directed** proposition is syntactically asserted), certain conditionals ("*Glaciers are melting because temperature rises*" in which both **rise** and **melt** predications are asserted), and several other constructions.

---
[6] Distinguished, in this case, by the copula **are** in the second sentence.

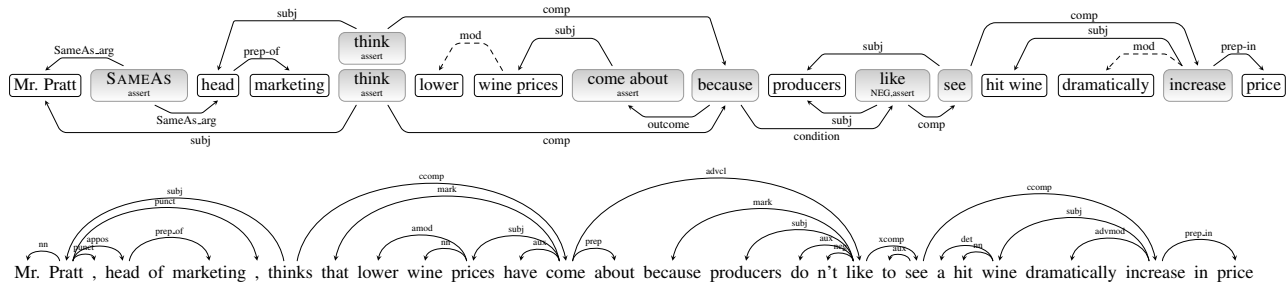

Figure 2: Our representation (top) versus Stanford-dependencies representation (bottom) of the sentence "Mr. Pratt, head of marketing, thinks that lower wine prices have come about because producers don't like to see a hit wine dramatically increase in price". See analysis in section 5

## 5 Concrete Example

We conclude the discussion of PROPS structures with a final example, taken from the PTB, which exhibits several of our transformations. Figure 2 compares PROPS with Stanford-dependency analysis for a typical WSJ sentence. The dependency representation contains 27 nodes (and therefore 27 edges), while our representation contains 17 nodes and 19 edges. A simple traversal from all predicate nodes in our graph yields the following propositions:

(1) lower wine prices have come about         [asserted]
(2) hit wine dramatically increase in price
(3) producers see (2)
(4) producers don't like (3)                  [asserted]
(5) Mr Pratt is the head of marketing         [asserted]
(6) (1) happens because of (4)
(7) Mr Pratt thinks that (6)                  [asserted]
(8) the head of marketing thinks that (6)     [asserted]

In contrast, extracting this set of propositions and their assertion status from the dependency tree requires a non-trivial amount of processing.

## 6 Annotated Corpus

In addition to the PROPS converter described in previous sections, we also created a large PROPS corpus, which extracts with high precision both the core structures described in Section 3, as well as the harder cases described in Section 4.[7]

To obtain this corpus we adapted the conversions described in Section 3 and ran them on full gold constituency trees, available from the Penn Tree Bank (PTB) (Bies et al., 1995). Using the PTB along with Propbank (which also annotated the WSJ), allows us to explicitly recover some of the phenomena which were only heuristically handled in the PROPS converter (Section 4). In order to recover argument propagation and conjunction handling we make use of the rich syntactic information available in the gold trees (including traces, empty elements and functional annotations). To distinguish raising-to-subject from control, we made use of Propbank annotations, as depicted in (Chrupała and van Genabith, 2007).

## 7 Evaluation

We evaluate the PROPS converter *intrinsically* (by matching its output to a gold standard) and *extrinsically* (by demonstrating the utility of the corresponding structure in a text-comprehension task).

### 7.1 Intrinsic Evaluation

In order to intrinsically test the performance of PROPS, we manually annotated a batch of 100 sentences of the WSJ, composed of the first 50 consecutive sentences of sections 2 (train) and 22 (dev). The annotation was performed by manually verifying and fixing the output of the automatic conversion of the WSJ, described in Section 6.

In order to test our *labeled attachment accuracy* we adapt the metric used in Ballesteros et al (2014) to account for PROPS' non 1-1 correspondence between nodes and sentence tokens (see Section 2). To this end, we define an edge as a triplet of (head-span, mod-span, label), where head-span and mod-span are each an ordered list of token indices that contribute to the node.[8] For an edge to be considered

---

[7] The corpus will be made available upon publication.

[8] For example, an edge of the form $((1, 2, 3), (4, 5), subj)$ indicates that a node over tokens 1-3 is the head of a node over tokens 4-5 with a label subj.

correct, both the nodes and the label should match.

More formally, given a gold reference graph $g$, its approximation $\hat{g}$, and the gold standard corpus $GOLD$ (a list of reference graphs), we take the list of edges $E(g) = \{(i_1, ..., i_k), (j_1, ..., j_l), label\}$, and calculate precision and recall metrics between $E(g)$ and $E(\hat{g})$, in the following manner:

$$P_{las} = \frac{\Sigma_{g \in GOLD}|E(g) \cap E(\hat{g})|}{\Sigma_{g \in GOLD}|E(\hat{g})|} \quad (1)$$

$$R_{las} = \frac{\Sigma_{g \in GOLD}|E(g) \cap E(\hat{g})|}{\Sigma_{g \in GOLD}|E(g)|} \quad (2)$$

We take a similar approach for computing *feature accuracy*.[9] Given the list of features $F(g) = \{(i_1, ..., i_k), (key, value)\}$ we calculate feature precision and recall metrics between $F(g)$ and $F(\hat{g})$:

$$P_{feat} = \frac{\Sigma_{g \in GOLD}|F(g) \cap F(\hat{g})|}{\Sigma_{g \in GOLD}|F(\hat{g})|} \quad (3)$$

$$R_{feat} = \frac{\Sigma_{g \in GOLD}|F(g) \cap F(\hat{g})|}{\Sigma_{g \in GOLD}|F(g)|} \quad (4)$$

**Results** We tested both the WSJ conversion (against the gold standard), as well as our automatic tool (against the gold standard and against the entire WSJ conversion). The results are shown in table 2.

We find that the WSJ conversion performs with 91% $F1$ score of for labeled attachment, and 96% $F1$ score for feature computation. Note that our evaluation metric is rather harsh, as a small node segmentation error will penalize both in terms of recall and precision all of the edges that include the incorrectly segmented nodes.[10]

Further manual inspection of the errors revealed that the vast majority ($> 80\%$) stem either from errors in the underlying annotations, or disagreement about annotation standards (e.g. PropBank marks "business" in "*the business grew*" as an object, while we treat it as a subject). This high accuracy of the automatic conversion, which stems from the use of the manual annotations, warrants its use as a large

---
[9] To be clear, a sentence fragment such as "*did not walk*" will result in a node "walk" with features tense=past, negated=True, covering three sentence tokens.

[10] For example, failing to combine "*New England Journal of Medicine*" into a single node in "*results appear in today's New England Journal of Medicine*" is penalized with 6 edge errors.

|  | Feature Computation | | | Modified LAS | | |
|---|---|---|---|---|---|---|
|  | P | R | F1 | P | R | F1 |
| WSJ | .95 | .97 | .96 | .9 | .92 | .91 |
| PROPS | .88(.88) | .89(.84) | .89(.86) | .83(.8) | .81(.81) | .82(.8) |

Table 2: Conversion accuracy, WSJ is compared against gold standard, PROPS against the gold standard and WSJ (in parentheses).

reference corpus for the development and evaluation of future parsers for PROPS.

Additionally, we test the accuracy of converting PROPS over automatic dependency trees derived by running the Stanford converter on top of the Berkeley parser (Petrov and Klein, 2007). While the PROPS parser is only capable of producing an approximation of the complete representation and is bounded by the accuracy of the underlying dependency parser (82%$F1$ for labeled attachment and 89%$F1$ for feature computation on the gold standard), it does succeed in recovering the major parts of the structure, as we demonstrate in the following section.

### 7.2 Extrinsic Evaluation

To extrinsically evaluate our automatic converter we use the MCTest corpus for machine comprehension (Richardson et al., 2013), composed of 500 short stories, each followed by 4 multiple choice questions. The MCTest comprehension task does not require extensive world knowledge. This makes it ideal for testing underlying representations, as performance will mostly depend on accuracy and informativeness of the representation. We focus our tests on questions which are marked in the corpus as answerable from a single sentence in the story (905 questions followed by 3620 candidate answers).

Our goal in this evaluation is to exemplify the usability of PROPS in an out-of-the-box application. We therefore focus on a simple format-independent algorithm, which allows us to control our evaluation around the underlying representation. The intention is to emphasize the attractive features of PROPS for semantic applications, rather than providing an elaborate, state-of-the-art text comprehension algorithm.

Richardson et al (2013) introduce a lexical matching algorithm, which we adapt to use either dependency or PROPS structures. They convert a question and a possible answer into a candidate answer (CA) (e.g., "*Billy is the name of the boy*" is obtained from the question/answer pair "what is the name of the boy?"/"Billy"), counts lexical matches between CA and a sliding window over the story, and returns the answer with the maximal number of matches.

| Method | Correct |
|---|---|
| PROPS | **66.34%** |
| dependencies | 64.58% |
| lexical | 60.44% |

Table 3: Results on MCTest corpus

Our adaptation works by counting matches of *representation-units* instead of words: The CA is compared against each story sentence (instead of a sliding window), and instead of counting lexical matches counts the number of identical representation-units in the CA and the sentence.[11] Table 3 shows the adaptation is effective: using dependencies improve over the lexical baseline reported in the MCTest paper, and using PROPS further improves results over dependency trees.

PROPS outperforms dependencies in answering questions involving phenomena addressed by our framework (e.g., in appositives: "*John and his best friend, Rick, shared their love for peaches.*" to answer: "What did John and Rick both love?", or adjectival complements as in "*she was wearing a costume that looked like a kitten*" to answer the question "What did Tessa's costume look like?").

## 8 Related Work

Abstracting away from syntax and providing a more semantically-oriented sentence representations has been the focus of several research efforts. In terms of the produced outputs, perhaps the most similar to ours is UCCA (Abend and Rappoport, 2013). They proposed a new representation scheme geared towards semantic applications and a small accompanying gold standard corpus. However, UCCA differs in their main motivations. Namely they aim to be a cognitive interlingua representation. The design choices of UCCA seem to focus on descriptiveness and universality rather than immediate use, resulting in a cognitively-motivated label-set that, for example, does not distinguish between subjects and objects giving both a "participant" status. Finally, to the best of our knowledge there is no automated system capable of inferring UCCA structures from text, while we provide a concrete implementation and demonstrate its effectiveness.

Semantic Role Labeling (SRL) (Carreras and Màrquez, 2005b) is also centered around proposition structures, with the goal of identifying predicates and linking each predicate to its semantic arguments (agent, patient, etc.) and adjuncts (locative, temporal, etc.). While SRL does not cover the full scope of propositions and unifications handled by PROPS, for the propositions it does cover it tackles a challenging semantic task that PROPS does not attempt to address, namely assigning arguments with *semantic* roles and grounding predicates to a semantic lexicon. PROPS graphs may be extended with SRL annotations, which can be incorporated on top of PROPS structures by adding semantic-role labels to predicate-argument relations.

Other notable semantic representations include GMB (Basile et al., 2012), UNL (Uchida and Zhu, 2001) and recently AMR (Banarescu et al., 2013). While AMR's rooted tree cannot capture independent propositions within a single sentence, it otherwise subsumes PROPS and SRL, aiming to provide the complete semantic structure of the sentence (along with inner structure and inter-relations). Producing accurate AMR structures requires in-depth semantic analysis, and it is currently produced with relatively low accuracy (Flanigan et al., 2014; Wang et al., 2015; Pust et al., 2015; Artzi et al., 2015). In contrast, PROPS is designed to be relatively easy to induce. Generally, PROPS can be viewed as a step between syntax and more semantic representations, providing a strong foundation on top of which more elaborate semantic annotations can be added.

## 9 Conclusion

We presented PROPS – a large set of linguistically motivated conversions. PROPS allows semantic applications to easily explore a broad range of the proposition structure, oblivious to the way it was expressed in the surface syntactic form. In addition, we produced an automatic high accuracy conversion of the WSJ, and showed the out-of-the-box applicability of the PROPS converter.

---

[11] Both PROPS and Stanford-dependency trees were obtained by using Berkeley Parser (Petrov and Klein, 2007) as the base parser. In both cases the representation-units are (source-word(s), label, target-word(s)) triplets.